\documentstyle[aclap]{article}

\newcounter{HoldEx}                  
\newcounter{SaveEx}

\newtheorem{Exmpl}{Example}

\newcommand{\startx} 
   {\par\noindent
    \begin{minipage}[t]{5.5in}
    \vspace*{0.25ex}
    \begin{Exmpl}
    \rm \ \\
    \makebox[.1in]{}
    \begin{minipage}[t]{5in}}

\newcommand{\stopx}  
    {\end{minipage}
     \end{Exmpl}
     \vspace*{0.1ex}
     \end{minipage}\\}

\include{psfig}

\title{\vspace{-0.5in}Expectations in Incremental Discourse Processing}

\author{Dan Cristea \\
Faculty~of Computer Science \\University ``A.I. Cuza''\\
16, Berthelot Street\\ 6600 - Iasi, Romania \\
{\tt dcristea@infoiasi.ro}\And
Bonnie Webber \\
Dept. of Computer \& Information Science \\
University of Pennsylvania \\
200 South 33rd Street \\
Philadelphia PA 19104-6389 USA \\
{\tt bonnie@central.cis.upenn.edu}
}
\begin{document}
\maketitle
\bibliographystyle{acl}
\vspace{-0.5in}
\begin{abstract}
The way in which discourse features express connections
back to the previous discourse has been described in the literature
in terms of {\em adjoining} at the {\em right frontier} of discourse
structure. But this does not allow for discourse features that express
{\em expectations} about
what is to come in the subsequent discourse.  After characterizing these
expectations and their distribution in text, we show how an approach
that makes use of {\em substitution} as well as {\em adjoining} on a
suitably defined right frontier, can be used to both process expectations
and constrain discouse processing in general.
\end{abstract}

\section{Introduction}

Discourse processing subsumes several distinguishable but interlinked
processes. These include
reference and ellipsis resolution, inference (e.g., inferential
processes associated with focus particles such as, in English, ``even''
and ``only''), and identification of those structures underlying a
discourse that are associated with coherence relations between its units.
In the course of developing an incremental approach to the
latter, we noticed a variety of constructions in discourse that
raise {\em expectations} about its
future structural features. We found that we could represent
such expectations by adopting a lexical variant of TAG -- LTAG
\cite{schabes90} -- and using its {\em substitution} operation
as a complement to {\em adjoining}. Perhaps more interesting was that
these expectations appeared to constrain the subsequent discourse until
they were resolved. This we found we could model in terms of constraints on
adjoining and substitution with respect to a suitably defined {\em Right
Frontier}. This short paper focuesses on the phenomenon of these
expectations in discourse and their expression in a discourse-level
LTAG. We conclude the paper with some thoughts on incremental discourse
processing in light of these expectations.

The following examples illustrate the creation of expectations through
discourse markers:
\startx
a. On the one hand, John is very generous.\\
b. On the other, he is extremely difficult to find.
\stopx
\setcounter{HoldEx}{\value{Exmpl}}
\startx
a. On the one hand, John is very generous.\\
b. On the other, suppose you needed some money.\\
c. You'd see that he's very difficult to find.
\stopx
\setcounter{SaveEx}{\value{Exmpl}}
\startx
a. On the one hand, John is very generous.\\
b. For example, suppose you needed some money.\\
c. You would just have to ask him for it.\\
b. On the other hand, he is very difficult to find.
\stopx
Example~\theHoldEx~ illustrates the expectation that, following a
clause marked ``on the one hand'', the discourse will express a
constrasting situation (here marked by
``on the other'').
Examples~\theSaveEx~ and \theExmpl~ illustrate that such
an expectation need not be satisfied immediately by the next
clause: In Example~\theSaveEx, clause (b) {\em partially} resolves the
expectation set up in (a), but introduces an expectation
that the subsequent discourse will indicate what happens in such cases.
That expectation is then
resolved in clause
(c). In Example~\theExmpl, the next two clauses do nothing to satisfy the
expectation raised in clause (a): rather,
they give evidence for the claim made in (a). The expectation raised
in (a) is not resolved until clause (d).

These examples show expectations raised by sentential adverbs
and the imperative use of the verb ``suppose''.
\setcounter{HoldEx}{\value{Exmpl}}
Subordinate conjunctions (e.g., ``just as'', ``although'', ``when'', etc.)
can lead to similar expectations when they appear in a preposed
subordinate clause - eg.
\startx
a. Although John is very generous,\\
b. if you should need some money,\\
c. you'd see that he's difficult to find.
\stopx
As in Example~\theSaveEx, clause~\theExmpl (a) raises the expectation
of learning what is nevertheless the case. Clause~\theExmpl (b)
partially satisfies that expectation by raising a hypothetical
situation, along with the expectation of learning what is true in such
a situation. This latter expectation is then satisfied in clause~\theExmpl (c).

In summary, these expectations can be characterized as follows:
(1) once raised, an expectation must be resolved, but its resolvant can
be a clause that raises its own expectations; (2) a clause
raising an expectation can itelf be elaborated
before that expectation is resolved, including elaboration by clauses
that raise their own expectations; and (3) the most deeply ``embedded''
expectations must always be resolved first.

Now these are very likely not the only kinds of expectations to be
found in discourse: Whenever events or behavior follow fairly regular
patterns over time, observers develop expectations about what will come
next or at least eventually. For example, a dialogue model may
embody the expectation that a {\em suggestion} made by one dialogue
participant would eventually be followed by an explicit or implicit
{\em rejection}, {\em acceptance} or {\em tabling} by the other.
Other dialogue actions such as {\em clarifications} or
{\em justifications} may intervene, but there is a sense of an expectation
being resolved when the {\em suggestion} is responded to.

Here we are focussed on discourse at the level of individual monologue
or turn within a larger discourse: what we show is that discourse
manifests certain forward-looking patterns that have similar constraints
to those of sentence-level syntax and can be handled by similar means.
One possible reason that these particualr kinds of expressions may
not have been noticed before is that in non-incremental approaches
to discourse processing \cite{mt88,marcu96}, they don't stand out as
obviously different.

The labels for discourse coherence relations used here are similar
to those of RST \cite{mt88}, but for simplicity, are treated as binary.
Since any multi-branching tree can be
converted to a binary tree, no representational power is lost.
In doing this, we follow several recent converging computational
approaches to discourse analysis, which are also couched in binary terms 
\cite{gardent97,marcu96,pol-van96,schilder97,vandenberg96}.

Implicit in our discussion is the view that in processing a discourse
incrementally, its semantics and pragmatics are computed compositionally
from the structure reflected in the coherence relations between its units.
In the figures presented here, non-terminal nodes in a discourse structure
are labelled with coherence relations merely to indicate the functions
that project appropriate content, beliefs
and other side effects into the recipient's discourse model.
This view is, we believe, consistent with the more detailed
formal interfaces to discourse semantics/pragmatics presented in
\cite{gardent97,schilder97,vandenberg96}, and also allows for multiple
discourse relations (intentional and informational) to hold between
discourse units \cite{mp92,mm95,mm96} and contribute to the
semantic/pragmatics effects on the recipient's discourse model.

\section{Expectations in Corpora}

The examples given in the Introduction were all ``minimal pairs'' created
to illustrate the relevant phenomenon as succinctly as possible. Empirical
questions thus include: (1) the range of
lexico-syntactic constructions that raise expectations with the
specific properties mentioned above; (2) the frequency of expectation-raising
constructions in text; (3) the frequency with which expectations are satisfied
immediately, as opposed to being delayed by material that elaborates the
unit raising the expectation; (4) the frequency of embedded expectations;
and (5) features that provide evidence for an expectation being satisfied.

While we do not have answers to all these questions, a very preliminary
analysis of the Brown Corpus, a corpus of approximately 1600 email messages,
and a short Romanian text by T. Vianu (approx. 5000 words)
has yielded some interesting results.

First, reviewing the 270 constructions that Knott has identified as potential
cue phrases in the Brown Corpus \footnote{Personal communication, but
also see \cite{knott96}}, one finds 15 adverbial phrases
(such as ``initially'', ``at first'', ``to start with'', etc.)
whose presence in a clause would lead to an expectation being raised.
All left-extraposed clauses in English raise expectations (as in
Example~\theExmpl) so all the subordinate conjunctions in Knott's
list would be included as well. Outside of cue phrases, we have identified
imperative forms of ``suppose'' and
``consider'' as raising expectations, but currently lack a more
systematic procedure for identifying expectation-raising constructions
in text than hand-combing text for them.

With respect to how often expectation-raising constructions appear in
text, we have Brown Corpus data on two specific types -- imperative
``suppose'' and adverbial ``on the one hand'' --
as well as a detailed analysis of the Romanian text by Vianu mentioned
earlier.

There are approximately 54K sentences in the Brown Corpus. Of these,
37 contain imperative ``suppose'' or ``let us suppose''. Twelve of these
correspond to ``what if'' questions or negotiation moves
which do not raise expectations:
\begin{quote}
  Suppose -- just suppose this guy was really what he
said he was!  A retired professional killer If he was just a nut,
no harm was done.  But if he was the real thing, he could do
something about Lolly. ({\em cl23})

Alec leaned on the desk, holding
the clerk's eyes with his.  ``Suppose you tell me the real reason'',
he drawled.  ``There might be a story in it''. ({\em cl21})
\end{quote}
The remaining 25 sentences constitute
only about 0.05\% of the Brown Corpus. Of these, 22 have their expectations
satisfied immediately (88\%) -- for example,
\begin{quote}
Suppose John Jones,
who, for 1960, filed on the basis of a calendar year, died June 20,
1961.  His return for the period January 1 to June 20, 1961, is due
April 16, 1962.
\end{quote}
One is followed by a single sentence elaborating the original supposition
(also flagged by ``suppose'') --
\begin{quote}
``Suppose it was not us that killed these aliens.  Suppose it
is something right on the planet, native to it.  I just hope it
doesn't work on Earthmen too.  These critters went real sudden''.
({\em cm04})
\end{quote}
while the remaining two contain multi-sentence elaborations of the
original supposition. None of the examples in the Brown Corpus contains
an embedded expectation.

The adverbial ``on the one hand'' is used to pose a
contrast either phrasally --
\begin{quote}
Both plans also prohibited common directors, officers, or employees between
Du Pont, Christiana, and Delaware, on the one hand, and General Motors
on the other. ({\em ch16})

You couldn't on the one hand
decry the arts and at the same time practice them, could you?  ({\em ck08})
\end{quote}
or clausally. It is only the latter that are of interest from the point
of discourse expectations.

The Brown Corpus contains only 7 examples of adverbial ``on the one hand''.
In three cases, the expectation is satisfied immediately by a clause
cued by ``but'' or ``or'' -- e.g.
\begin{quote}
On the one hand, the Public
Health Service declared as recently as October 26 that present
radiation levels resulting from the Soviet shots ``do not warrant
undue public concern'' or any action to limit the intake of
radioactive substances by individuals or large population groups
anywhere in the Aj.  But the PHS conceded that the new radioactive
particles ``will add to the risk of genetic effects in succeeding
generations, and possibly to the risk of health damage to some people
in the United States''.({\em cb21})
\end{quote}
In the remaining four cases, satisfaction of the expectation (the ``target''
contrast item) is delayed by 2-3 sentences elaborating the ``source''
contrast item -- e.g.
\begin{quote}
Brooklyn College students have an ambivalent attitude toward
their school.  On the one hand, there is a sense of not
having moved beyond the ambiance of their high school.  This is
particularly acute for those who attended Midwood High School directly
across the street from Brooklyn College.  They have a sense of
marginality at being denied that special badge of status, the
out-of-town school.  At the same time, there is a good deal of
self-congratulation at attending a good college \ldots ({\em cf25})
\end{quote}
In these cases, the target contrast item is cued by ``on the other
hand'' in three cases and ``at the same time'' in the case given above.
Again, none of the examples contains an embedded expectation.

(The much smaller email corpus contained six examples of clausal ``on the
one hand'', with the target contrast cued by ``on the other hand'',``on the other'' or ``at the other extreme''. In one case, there was
no explicit target contrast and the expectation raised by ``on the
one hand'' was never satisfied. We will continue to monitor for such
examples.)

Before concluding with a close analysis of the Romanian text, we should note
that in both the Brown Corpus and the email corpus, clausal adverbial
``on the other hand'' occurs more frequently {\em without} an
expectation-raising ``on the one hand'' than it does {\em with} one.
(Our attention was
called to this by a frequency analysis of potential cue phrase
instances in the Brown Corpus compiled for us by Alistair Knott and
Andrei Mikheev, HCRC, University of Edinburgh.) We found 53 instances
of clausal ``on the other hand'' occuring without an explicit source
contrast cued earlier. Although one can only speculate now
on the reason for this phenomenon, it does make a difference to
incremental analysis, as we try to show in Section~\ref{ex:sec}.

The Romanian text that has been closely analysed for explicit
expectation-raising constructions is T. Vianu's {\em Aesthetics}.
It contains 5160 words and 382 discourse units (primarily clauses).
Counting preposed gerunds as raising expectations as well as counting
the constructions noted previously, 39 instances of expectation-raising
discourse units were identified (10.2\%). In 11 of these cases, 1-16
discourse units intervened before the raised expectation was satisfied.
One example follows:
\begin{quote}
Dar de\c si trebuie s\u a-l parcurgem \^\i n \^\i ntregime,
pentru a orienta cercetarea
este nevoie s\u a \^\i ncerc\u am \^\i nc\u a de pe acum o precizare a obiectului lui.

({\em But although we must cover it entirely, in order to guide the research 
we need to try already an explanation of its subject matter.})
\end{quote}

\section{A Grammar for Discourse}

The intuitive appeal of Tree-adjoining Grammar (TAG) \cite{joshi87} for
discourse processing \cite{gardent97,pol-van96,schilder97,vandenberg96,webb91}
follows from the fact that TAG's {\em adjoining} operation allows one to
directly analyse the current discourse unit as a {\em sister} to previous
discourse material that it stands in a particular relation to. The new
intuition presented here -- that expectations convey a
dependency between the current discourse unit and future discourse
material, a dependency that can be ``stretched'' long-distance by
intervening material -- more  fully exploits TAG's ability to express
dependencies. By expressing in an elementary
TAG tree, a dependency betwen the current discourse unit and future
discourse material and using {\em substitution} \cite{schabes90}
when the expected material is found, our TAG-based approach to discourse
processing allows expectations to be both raised and resolved.

\subsection{Categories and Operations}

\begin{figure*}
\centerline{\psfig{figure=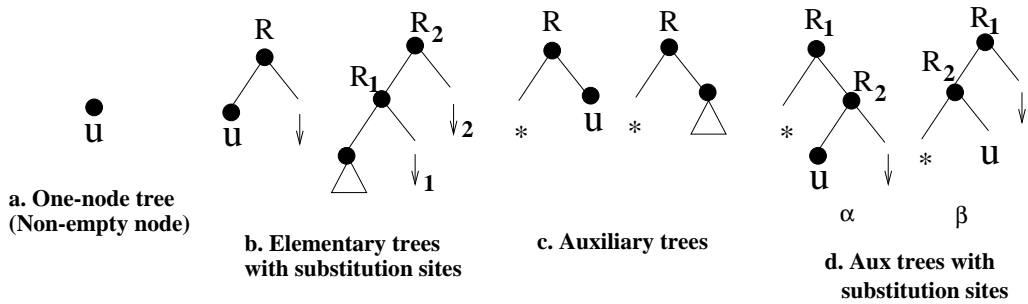}}
\label{gramcat:fig}
\caption{Grammatical Categories. (* marks the foot of an auxiliary tree,
and $\downarrow$, a substitution site.)}
\end{figure*}
The categories of our TAG-based approach consist of nodes and binary trees.
We follow \cite{gardent97} in associating nodes with feature structures
that may hold various sorts of information, including information about
the semantic interpretations projected through the nodes, constraints on
the specific operations a node may participate in, etc.
A non-terminal node represents a discourse relation holding
between its two daughter nodes. A terminal node can be either
{\em non-empty} (Figure~1a), corresponding to a basic discourse
unit (usually a clause), or {\em empty}. A node is ``empty'' only
in not having an associated discourse unit or relation: it
can still have an associated feature structure. Empty nodes play a
role in adjoining and substitution, as explained below, and 
hence in building the derived binary tree that represents
the structure of the discourse.

Adjoining adds to the discourse structure an {\em auxiliary tree} consisting
of a root labelled with a discourse relation, an empty {\em foot} node
(labelled *), and at least one non-empty node (Figures~1c and 1d). In our
approach, the {\em foot node} of an auxiliary tree must be its leftmost
terminal because all adjoining operations take place on a suitably
defined {\em right frontier} (i.e., the path from the root of a tree to
its rightmost leaf node) -- such that all newly introduced
material lies to the right of the adjunction site.
(This is discussed in Section~\ref{constr:sec} in more detail.)
Adjoining corresponds to identifying a discourse
relation between the new material and material in the
previous discourse that is still open for elaboration.

Figure~2(a) illustrates adjoining midway down the RF of tree $\alpha$,
while Figure~2(b) illustrates adjoining at the root of $\alpha$'s RF.
Figure~2(c) shows adjoining at the ``degenerate'' case of
a tree that consists only of its root. Figure~2(d) will be
explained shortly.

Substitution unifies the root of a {\em substitution structure} with an
empty node in the discourse tree that serves as a {\em substitution site}.
We currently use two kinds
of substitution structures: {\em non-empty nodes} (Figure~1a) and
{\em elementary trees with substitution sites} (Figure~1b). The latter
are one way by which a substitution site may be introduced
into a tree.  As will be argued shortly, substitution sites
can only appear on the {\em right} of an elementary tree,
although any number of them may appear there (Figure~1b). Figure~2(e)
illustrates substitution of a non-empty node at $\downarrow$,
and Figure~2(f) illustrates substitution of an elementary tree with its
own substitution site at $\downarrow_1$

\begin{figure*}
\centerline{\psfig{figure=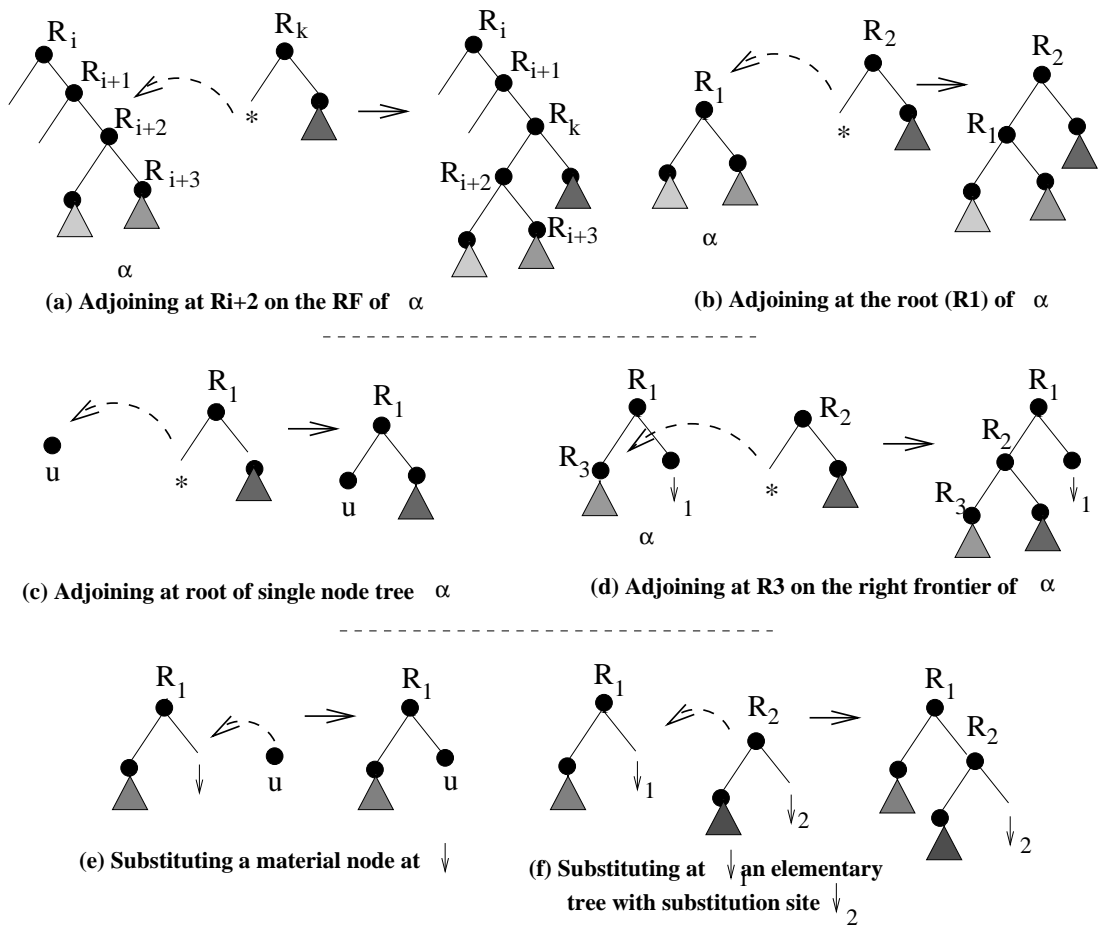}}
\label{adj_sub:fig}
\caption{Examples of Adjoining and Substitution}
\end{figure*}

Since in a clause with two discourse
markers (as in Example~\theHoldEx b) one may look backwards
(``for example'') while the other looks forwards (``suppose''),
we also need a way of introducing expectations in the context of
adjoining. This we do by allowing an {\em auxiliary tree} to contain
substitution sites (Figure~1d) which, as above, can only appear on
its right.\footnote{We currently have no linguistic evidence for
the structure labelled $\beta$ in Figure~1d, but are open to its
possibility.} Another term we use for auxiliary trees is
{\em adjoining structures}.

\subsection{Constraints}
\label{constr:sec}

Earlier we noted that in a discourse structure with no substitution
sites, adjoining is limited to the {\em right frontier} (RF). This is
true of all existing TAG-based approaches to discourse
processing \cite{gardent97,hinpol86,pol-van96,schilder97,webb91},
whose structures correspond to trees that lack substitution
sites. One reason for this RF restriction is to maintain a strict
correspondence between a left-to-right reading of the terminal nodes of
a discourse structure and the text it analyses - i.e.,
\begin{quote}
{\bf Principle of Sequentiality}: A left-to-right reading of the
{\em terminal frontier} of the tree associated with a discourse must
correspond to the span of text it analyses in that same left-to-right order.
\end{quote}
Formal proof that this principle leads to the
restriction of adjoining to the right frontier is given in \cite{cw97b}.

The Principle of Sequentiality leads to additional
constraints on where adjoining and substitution can occur in trees
with substitution sites. Consider the tree in Figure~3(i), which has
two such sites, and an adjoining operation on the right frontier at
node R$_j$ or above. Figure~3(ii) shows that this would introduce a
non-empty node (u$_k$) above and to the right of the substitution
sites. This would mean that later substitution at either of them
would lead to a violation of the Principle of
Sequentiality, since the newly substituted node u$_{k+l}$ would then
appear to the
left of u$_k$ in the terminal frontier, but to the right of it in
the original discourse. Adjoining at any node above R$_{j+2}$ --
the left sister of the most deeply embedded substitution site --
leads to the same problem (Figure~3iii). Thus in a tree with
substitution sites, adjoining must be limited to {\em nodes on the path
from the left sister of the most embedded site to that sister's rightmost
descendent}. But this is just a {\em right frontier} (RF) rooted at that left
sister. Thus, adjoining is always limited to a RF: the presence
of a substitution site just changes what node that RF is rooted at. We
can call a RF rooted at the left sister of the most embedded substitution
site, the {\em inner} right frontier or ``inner\_RF''. (In Figure~3(i),
the inner\_RF is indicated by a dashed arrow.) In contrast,
we sometimes call the RF of a tree without substitution sites, the {\em
outer} right frontier or ``outer\_RF''. Figure~2(d) illustrates
adjoining on the inner\_RF of $\alpha$, a tree with a substitution
site labelled $\downarrow_1$.
\begin{figure*}
\centerline{\psfig{figure=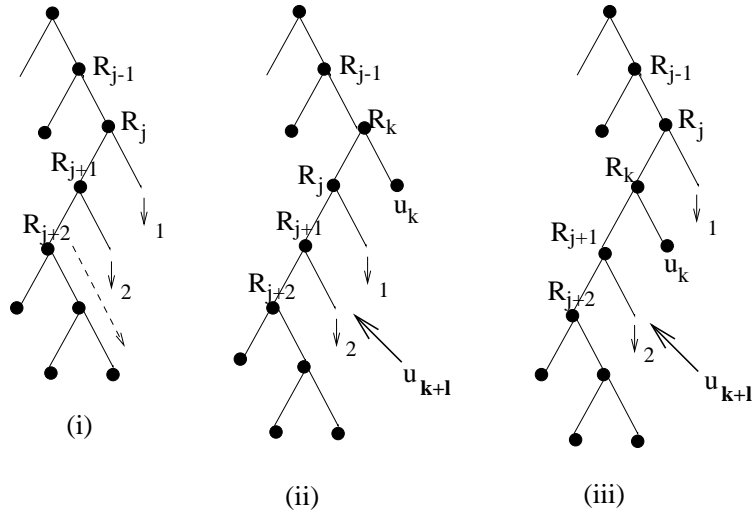}}
\label{gapAdjunct:fig}
\caption{Adjoining is constrained to nodes the inner\_RF, indicated
by the dashed arrow.}
\end{figure*}

Another consequence of the Principle of Sequentiality is that the only
node at which substitution is allowed in a tree with
substitution sites is at the most embedded one. Any other substitution
would violate the principle. (Formal proof of these claims are given in
\cite{cw97b}.

\subsection{Examples}
\label{ex:sec}

Because we have not yet implemented a parser that embodies the ideas
presented so far, we give here an idealized
analysis of Examples~\theSaveEx~ and \theHoldEx, to show
how an ideal incremental monotonic algorithm that admitted
expectations would work.
\begin{figure*}
\centerline{\psfig{figure=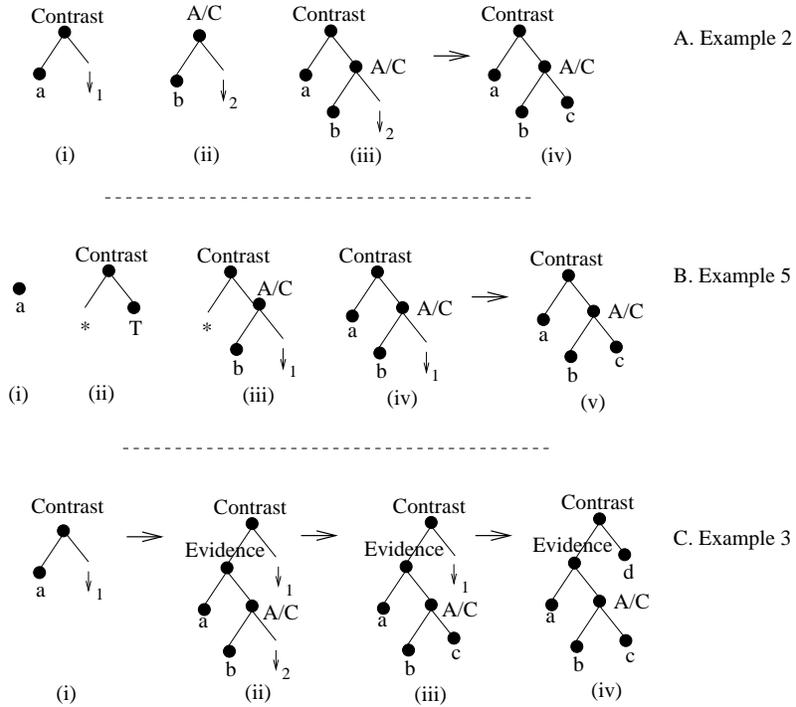}}
\label{ex2:fig}
\caption{Analyses of Examples~\theSaveEx, \theHoldEx~ and \theExmpl.}
\end{figure*}

Figure~4A illustrates the incremental analysis of Example~\theSaveEx.
Figure~4A(i) shows the elementary tree corresponding
to sentence~\theSaveEx a (``On the one hand \ldots''): the interpretation of
``John is very generous'' corresponds to the left daughter labelled
``a''. The adverbial ``On the one hand'' is taken as signalling a coherence
relation of {\bf Contrast} with something expected later in the discourse.

In sentence~\theSaveEx b (``On the other hand, suppose \ldots''), the
adverbial ``On the other hand'' signals the
expected contrast item. Because it is already expected, the adverbial
does not lead to the creation of a separate elementary tree (but see the
next example).
The imperative verb ``suppose'', however, signals a coherence relation of
{\bf antecedent/consequent} (A/C) with a consequence expected later in the
discourse. The elementary tree corresponding to ``suppose \ldots''
is shown in Figure~4A(ii), with the interpretation of ``you need money''
corresponding to the left daughter labelled ``b''. Figure~4A(iii) shows
this elementary tree {\em substituted} at $\downarrow_1$,
satisfying that expectation.
Figure~4A(iv) shows the interpretation of sentence~\theSaveEx c (``You'd
see he's very difficult to find'')
substituted at $\downarrow_2$, satisfying that remaining expectation.

Before moving on to Example 3, notice that if Sentence~\theSaveEx a were
not explicitly cued with ``On the other hand'',
the analysis would proceed somewhat differently.
\startx
a. John is very generous.\\
b. On the other hand, suppose you needed money.\\
c. You'd see that he's very difficult to find.
\stopx
Here, the interpretation of sentence~\theExmpl (a) would correspond to the
degenerate case of a tree consisting of a single non-empty node shown in
Figure~4B(i).
The contrast introduced by ``On the other hand'' in sentence~\theExmpl (b)
leads to the auxiliary tree shown in Figure~4B(ii),
where {\bf T} stands for the elementary tree corresponding to the
interpretation of ``suppose \ldots''. The entire structure associated
with sentence~\theExmpl (b) is shown in Figure~4B(iii).
This is adjoined to the single node tree in  Figure~4B(i), yielding
the tree shown in Figure~4B(iv). The analysis then continues
exactly as in that of Example~\theSaveEx~ above.

Moving on to Example~\theHoldEx, Figure~4C(i) shows the same elementary
tree as in Figure~4A(i) corresponding to clause~\theHoldEx a.
Next, Figure~4C(ii) shows the auxiliary tree with substitution site
$\downarrow_2$ corresponding
to clause~\theHoldEx b being adjoined as a sister to the interpretation
of clause~\theHoldEx a, as evidence for the claim made there.
The right daughter of the node labelled ``Evidence'' is, as in
Example~\theSaveEx b, an elementary tree expecting the consequence of the
supposition ``you need money''. Figure~4C(iii) shows the
interpretation of clause~\theHoldEx c substituted at $\downarrow_2$,
satisfying that expectation. Finally, Figure~4C(iv) shows the
interpretation of clause~\theHoldEx d substituted at $\downarrow_1$,
satisfying the remaining expectation.

\section{Sources of Uncertainty}
\label{uncertain:sec}

The idealized analysis presented above could lead to a simple
deterministic incremental algorithm, if there were no uncertainty due
to local or global ambiguity. But there is. We can identify
three separate sources of uncertainty that would affect
incremental processing according to the grammar just presented:
\begin{itemize}
\item the identity of the discourse relation that is meant to hold between
two discourse units;
\item the operation (adjoining or substitution) to be used in adding
one discourse unit onto another;
\item if that operation is adjoining, the site in the target unit at
which the operation should take place -- that is, the other argument
to the discourse relation associated with the root of the auxiliary tree.
\end{itemize}

It may not be obvious that there could be uncertainty as to whether
the current discourse unit satisfies an expectation
and therefore {\em substitutes} into the discourse structure,
or elaborates something in the previous discourse, and therefore
{\em adjoins} into it.\footnote{This is not the same as shift-reduce
uncertainty.} But the {\em evidence} clarifying this local ambiguity
may not be available until later in the discourse. In the
following variation of Example~4, the fact that
clause~(b) participates in elaborating the interpretation of clause~(a)
rather than in satisfying the expectation it raises (which it does in
Example~4) may not be unambiguously clear until the discourse marker
``for example'' in clause~(c) is processed.
\startx
a. Because John is such a generous man --\\
b. whenever he is asked for money,\\
c. he will give whatever he has, for example --\\
d. he deserves the ``Citizen of the Year'' award.
\stopx
The other point is that, even if a forward-looking cue phrase
signals only a substitution structure as in Figure~4A(i) and 4A(ii),
if there are no pending subsitution sites such as $\downarrow_1$ in 4A(i)
against which to unify such a structure, then the substitution structure
must be {\em coerced} to an auxiliary tree as in Figure~1d (with some
as yet unspecified cohesion relation) in order to adjoin it somewhere
in the current discourse structure.

\section{Speculations and Conclusions}

In this paper, we have focussed on discourse expectations
associated with forward-looking clausal connectives, sentential
adverbs and the imperative verbs (``suppose'' and ``consider'').
There is clearly more to be done, including a more complete
characterization of the phenomenon and development of an
incremental discourse processor based on the ideas presented above.
The latter would, we believe, have to be coupled with
incremental sentence-level processing. As the previous examples have shown,
the same phenomenon that occurs inter-sententially in Examples~1-3
occurs intra-sententially in Examples~4 and 6, suggesting that the two
processors may be based on identical principles. In addition,
carrying out
sentence-level processing in parallel with discourse processing and
allowing each to inform the other would allow co-reference interpretation
to follow from decisions about discourse relations and vice versa.

\section{Acknowledgements}
Support
for this work has come from the Department of Computer Science, Universiti
Sains Malaysia (Penang, Malaysia), the Department of Computer Science,
University ``A.I.Cuza'' (Iasi, Romania) and the Advanced Research
Project Agency (ARPA) under grant N6600194C6-043 and the Army Research
Organization (ARO) under grant DAAHO494GO426.
Thanks go to both the anonymous reviewers and the
following colleagues for their helpful comments: Michael Collins,
Claire Gardent, Udo Hahn, Joseph Rosenzweig, Donia Scott, Mark Steedman,
Matthew Stone, Michael Strube, and Michael Zock. Thanks also
to Alistair Knott and Andrei Mikheev
for giving us a rough count of cue phrases in the Brown Corpus.

\end{document}